\documentclass[sigconf,natbib=true]{acmart}
\usepackage{booktabs} %
\usepackage{graphicx}  %
\usepackage{multirow}
\usepackage{subfigure}
\usepackage{xspace}
\usepackage{colortbl}
\usepackage{xcolor}
\usepackage{pifont}
\usepackage{amsmath}
\usepackage{amsfonts}
\usepackage{resizegather}
\usepackage{algorithm}
\usepackage{algorithmic}

\newcommand{\aka}{\emph{a.k.a.,}\xspace}
\newcommand{\eg}{\emph{e.g.,}\xspace}

\newcommand{\ignore}[1]{}

\newcommand{\dubbelop}{$^{\blacktriangle}$}

\newcommand{\dubbelneer}{$^{\blacktriangledown}$}

\newcommand{\fullmodel}{\textbf{H}eterogeneous \textbf{H}istorical \textbf{R}eview aware \textbf{R}eview \textbf{S}ummarization Model\xspace}
\newcommand{\model}{HHRRS\xspace}

\AtBeginDocument{%
  }

\setcopyright{acmcopyright}
\copyrightyear{2018}
\acmYear{2018}
\acmDOI{XXXXXXX.XXXXXXX}

\acmConference[Conference acronym 'XX]{Make sure to enter the correct
  conference title from your rights confirmation emai}{June 03--05,
  2018}{Woodstock, NY}
\acmPrice{15.00}
\acmISBN{978-1-4503-XXXX-X/18/06}

\begin{document}

\title{Towards Personalized Review Summarization by Modeling Historical Reviews from Customer and Product Separately}

\author{Xin Cheng\textsuperscript{2}, Shen Gao\textsuperscript{1}, Yuchi Zhang\textsuperscript{3}, Yongliang Wang\textsuperscript{3}, \\ Xiuying Chen\textsuperscript{6}, Mingzhe Li\textsuperscript{3}, Dongyan Zhao\textsuperscript{2} and Rui Yan\textsuperscript{4,5}}
\authornote{Corresponding Author: Rui Yan (ruiyan@ruc.edu.cn) and Dongyan Zhao (zhaody@pku.edu.cn)}
\authornote{Xin Cheng and Shen Gao contribute equally to this paper. Ordering is decided by a coin flip.}
\affiliation{
  \institution{
    \textsuperscript{1} School of Computer Science and Technology, Shandong University \\
    \textsuperscript{2} Wangxuan Institute of Computer Technology, Peking University \\
    \textsuperscript{3} Ant Group \\
    \textsuperscript{4} Gaoling School of Artificial Intelligence, Renmin University of China \\
    \textsuperscript{5} Beijing Academy of Artificial Intelligence \\
    \textsuperscript{6} King Abdullah University of Science and Technology \\
  }
  \country{}
  \city{}
}
\email{chengxin1998@stu.pku.edu.cn, {shengao, li_mingzhe, zhaody}@pku.edu.cn, {yuchi.zyc, yongliang.wyl}@alibaba-inc.com}
\email{xiuying.chen@kaust.edu.sa, ruiyan@ruc.edu.cn}

\renewcommand{\shortauthors}{Xin Cheng, Shen Gao et al.}

\begin{abstract}
    Review summarization is a non-trivial task that aims to summarize the main idea of the product review in the E-commerce website.
    Different from the document summary which only needs to focus on the main facts described in the document, review summarization should not only summarize the main aspects mentioned in the review but also reflect the personal style of the review author.
    Although existing review summarization methods have incorporated the historical reviews of both customer and product, they usually simply concatenate and indiscriminately model this two heterogeneous information into a long sequence.
    Moreover, the rating information can also provide a high-level abstraction of customer preference, it has not been used by the majority of methods.
    In this paper, we propose the \fullmodel (\model) which separately models the two types of historical reviews with the rating information by a graph reasoning module with a contrastive loss.
    We employ a multi-task framework that conducts the review sentiment classification and summarization jointly.
    Extensive experiments on four benchmark datasets demonstrate the superiority of \model on both tasks.
\end{abstract}


\maketitle

\begin{figure}
    \centering
    \includegraphics[width=1.0\columnwidth]{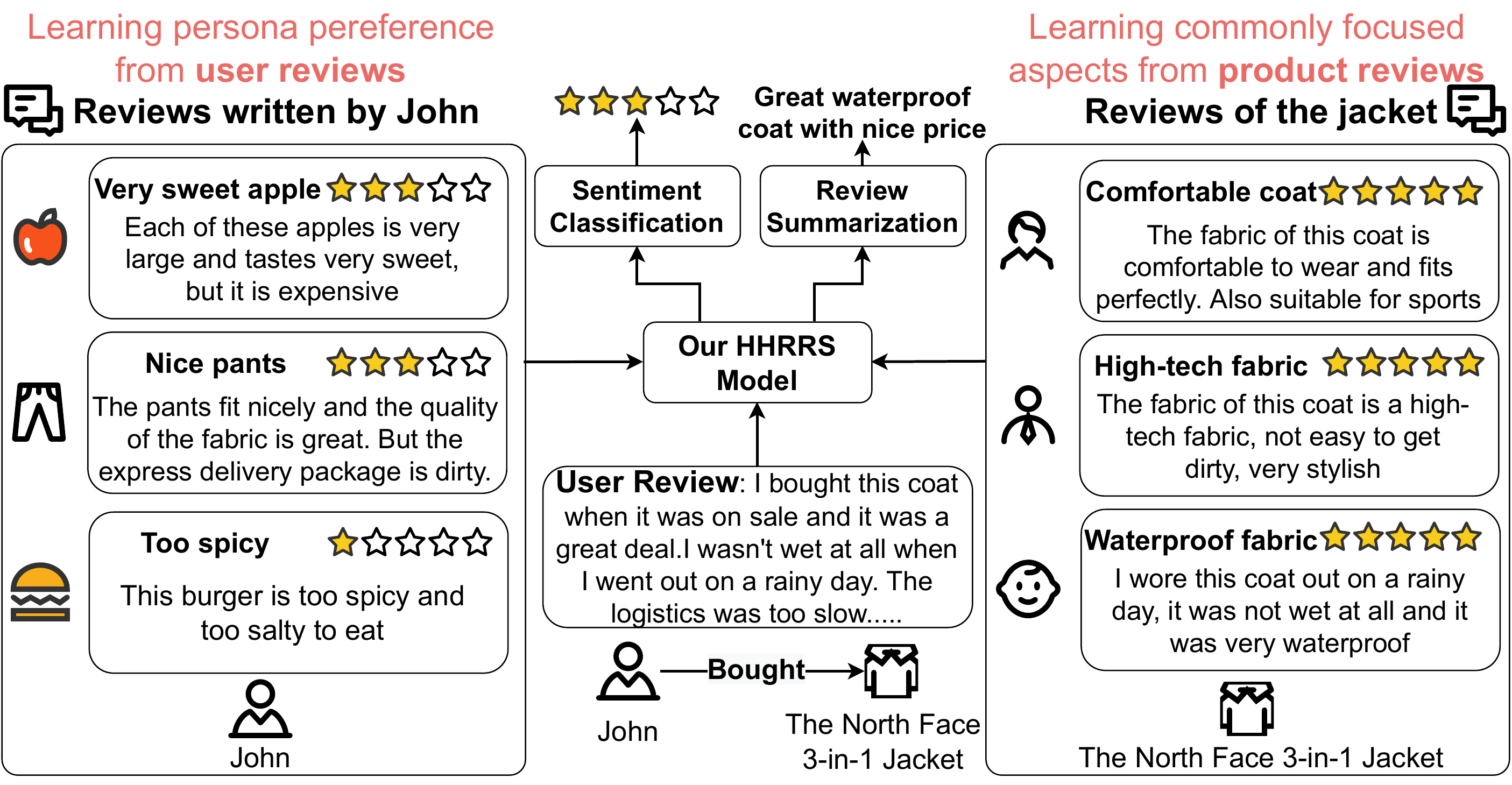}
    \caption{
        Example of incorporating historical reviews and rating for review summarization. 
        From the customer review, we can see that the customer is quite picky and always give borderline or negative scores. 
        From the product review, we can learn the commonly focused aspect (\eg the fabric of jacket).
    }
    \label{fig:intro}
\end{figure}

\section{Introduction}

Most of the E-commerce portals provide a review panel for customers who have already bought the product to write a review of their experience.
Many customers not only write a review but also give a short summary of the review, which can help other consumers to know the product better.

Different from other text summarization tasks, the product review summarization is highly personalized and product-centric~\cite{Li2019Towards,Amplayo2021AspectControllable}.
To be more specific, a good summary should (1) reflect the \textbf{persona writing preference} of the customer and (2) describe the \textbf{commonly focused aspects} of this product which are useful for future customers.
These two requirements can potentially be met by utilizing historical reviews, where the customer's historical reviews reflect the writing style, and the historical product reviews describe commonly focused aspects.
Following this direction, researchers proposed to incorporate the historical reviews~\cite{Xu2021Transformer,Liu2019Neural} for review summarization.
However, these existing methods usually mix the historical reviews of customers and products together by concatenating them into a long sequence.
Since we aim to learn the writing style from historical reviews of the customer and learn the main focused aspects of the product from the historical product reviews, these two reviews should play different roles in guiding the summarization process. 
Therefore, our first challenge is \textit{how to fully explore the two kinds of information from the two types of reviews and take advantage of their respective roles in generating summaries}.

In the meantime, the review rating can be seen as a \textbf{high-level abstraction of the review} which reflects the satisfaction of the customer with the product.
The customer rating reviews contain personal rating preferences, and the rating for the same product reflects the average user satisfaction with the product.
Figure~\ref{fig:intro} shows an example of using two types of historical reviews and ratings can capture the actual user preference.
Thus, modeling the historical review ratings can help the model understand the user's satisfaction with the product.
To the best of our knowledge, the rating information has not been explored in related works~\cite{Ma2018A,Chan2020A}.
Therefore, our second challenge is \textit{how to incorporate the rating of historical reviews to predict sentiment better and generate a personalized summary.}

To tackle these two challenges, in this paper, we propose a personalized review summarization model named \fullmodel (\model).
Different from previous methods, \model first (1) separately models the relationship between reviews of the customer and product by a graph reasoning model; (2) incorporates the rating information for the historical reviews.
By these two methodologies, our model can understand the customer persona writing style and the main focused aspects of the product better, and improve the performance of two tasks.
For the \textit{first challenge}, we construct two graphs for historical reviews of customer and product separately to capture the relationship and model the interaction between reviews.
Since the two types of review are similar in literal, to force the model to learn customer writing style from customer reviews and extract the salient product aspects from product reviews, we propose a contrastive learning module that prevents the graph module from learning the homogeneous information from customer reviews and product reviews.
And for the \textit{second challenge}, since the rating of review provides high-level information about the review, we employ the rating for the historical customer and product reviews to capture the rating preference of the customer and product respectively.

Previous studies~\cite{Ma2018A,Chan2020A} show that jointly training the review sentiment classification (\aka rating prediction) model with the summarization model can boost the performance of both tasks.
Motivated by these works, we first introduce \textit{historical} review ratings into the review sentiment classification task and propose a multi-task paradigm.
Finally, we generate a personalized summary by incorporating the historical reviews and input reviews with a graph-attention layer.
Experiments conducted on the benchmark datasets verify the effectiveness of our proposed model compared to the state-of-the-art baselines in sentiment classification and summarization tasks.

\noindent To sum up, our contributions can be summarized as follows:

$\bullet$ We propose to separately model the historical customer and product reviews to capture the personal style and commonly focused aspects of the product by a graph-based reasoning model.

$\bullet$ We incorporate the rating of historical reviews in the summarization process, which provide high-level information for the review summarization. %

$\bullet$ Experiments show the superiority of \model compared with state-of-the-art baselines on summarization and sentiment classification tasks.

\section{Related Work}\label{sec:related}

\subsection{Document Summarization}

Document summarization aims to produce a short summary that covers the main idea of the input document.
These methods can be classified into two categories: generative and extractive.
Extractive summarization methods select several salient sentences from the input document as the summary, while abstractive summarization methods write the summary from scratch.
In recent years, the pre-train language model (PLM)~\cite{Lewis2020BART,Devlin2019BERTPO,Liu2019RoBERTaAR} shows its great potential in language understanding and language generation tasks.
Many researchers employ the PLM to obtain the contextualized sentence representation which can help the extractive summarization model achieve better performance.
However, the extractive summarization methods usually produce a summary with redundant information and the summary is not coherent, since these methods simply concatenate several discontiguous sentences as a summary.
The abstractive summarization methods, especially based on the large-scale PLM, can generate a more fluent and condensed summary than the extractive-based methods.
From the experimental results on several benchmark document summarization datasets, we can find that the abstractive summarization methods outperform the extractive methods.
In this paper, we focus on the review summarization task which usually needs to incorporate contextual information to produce a better summary (\eg product information, customer persona), and it should describe the popular product aspects.
Thus, directly employing the document summarization methods on review cannot achieve good performance.

\subsection{Review Summarization}

Review summarization aims to produce a brief summary of the e-commerce product review.
Early review summarization methods are mostly based on extractive methods, which directly extract phrases and sentences from the original review as the summary.
\cite{Hu2004MiningAS} mine the features of the product from the customer review and identify whether the opinions are positive or negative.
\cite{Xiong2014EmpiricalAO} propose an unsupervised extractive review summarization method that exploits review helpfulness ratings.

For the abstractive methods, \cite{Chan2020A,Ma2018A} propose a multi-task framework to leverage the shared sentiment information in both review summarization and sentiment classification tasks.
\cite{Liu2019Neural,Xu2021Transformer} propose the transformer-based reasoning framework for the personalized review summarization model, which first concatenates the historical reviews of customer and product and feeds into the reasoning layer.

Although existing methods incorporate historical reviews, these methods simply concatenate all the reviews of customer and product and they cannot identify the different information from customer persona style and product aspects.
And most of the existing methods ignore the rating information of the historical reviews.
We compare the characteristics of several cutting-edge review summarization methods and our \model in Table~\ref{tab:xiaomi_comp}.

\begin{table}[t]
    \begin{center}
    \caption{Characteristics of different methods. We not only model the heterogeneity of historical reviews, but also combines the advantages of existing methods.}
    \label{tab:xiaomi_comp}
    \resizebox{1\columnwidth}{!}{
    \begin{tabular}{c|cccccc}
    \toprule
    & PGNet~\cite{See2017Get} & HSSC~\cite{Ma2018A} & DualView~\cite{Chan2020A} & TRNS~\cite{Xu2021Transformer} & \model \\ %
    \midrule
    Customer Reviews         & \ding{55} & \ding{55}  & \ding{55}  & \ding{51}  & \ding{51} \\
    Product Reviews          & \ding{55} & \ding{55}  & \ding{55}  & \ding{51}  & \ding{51} \\
    Heterogeneity Modeling      & \ding{55} & \ding{55}  & \ding{55}  & \ding{55}  & \ding{51} \\
    Review Relation Modeling & \ding{55} & \ding{55}  & \ding{55}  & \ding{55}  & \ding{51} \\
    \midrule
    Sentiment Classification & \ding{55} & \ding{51}  & \ding{51}  & \ding{55}  & \ding{51} \\
    Historical Sentiment    & \ding{55} & \ding{55}  & \ding{55}  & \ding{55}  & \ding{51} \\
    \bottomrule
    \end{tabular}
    }
    \end{center}
\end{table}

\section{Problem Formulation}\label{sec:problem-formulation}

Given an input review $r = \{r_1, \cdots, r_{L_r}\}$ with $L_r$ tokens which is written by customer $u$ for product $p$, our goal is to generate a summary $\hat{y} = \{\hat{y}_1, \cdots, \hat{y}_{L_y}\}$ with $L_y$ tokens.
To help the summarization model capture the customer style and preference and the common product aspects, we incorporate the historical reviews of customer $r^{u}$ and product $r^{p}$.
We use $r^{u,k} = \{r^{u,k}_1, \cdots, r^{u,k}_{L_r}\}$ to denote the $k$-th review of the same customer of review $r$, and $r^{p,k} = \{r^{p,k}_1, \cdots, r^{p,k}_{L_r}\}$ to denote the $k$-th review of the same product of review $r$.
Since we also use the sentiment classification as a multi-task, we use the rating $s^r$ for the input review $r$ and rating $s^{u}, s^{p}$ for historical reviews of customer and product respectively.
Finally, we use (1) the difference between generated summary $\hat{y}$ and the ground truth summary $y$ and (2) the difference between the predicted rating and the ground truth rating as the training objective.

\section{Preliminary}\label{sec:preliminary}

\subsection{Text Generation with Transformer}

Transformer~\cite{Vaswani2017Attention} is an encoder-decoder framework that captures the deep interaction between words in a sentence by using multi-head attention.
We start by introducing the encoder in Transformer.
It first projects the input text words into vector representation by an embedding matrix $e$ and then employs a multi-head self-attention mechanism.
We project the input embedding $e(r)$ into query, key, and value which are three dependent vector spaces:
\begin{equation}
    \begin{aligned}\label{equ:self-attn}
        & \operatorname{Attention}(r)= \\
        & \operatorname{Softmax}\left(\frac{(e(r) W^Q) (e(r) W^K)}{\sqrt{d}}\right) (e(r) W^V),
    \end{aligned}
\end{equation}
where $W^Q, W^K, W^V$ are all trainable parameters, $e(r)$ are the embeddings of each token in the review $r$, and $d$ is the dimension of embedding vector.
After interacting with other tokens in the input text, we apply the Feed-Forward Networks (FFN) on the output of Equation~\ref{equ:self-attn}:
\begin{align}\label{equ:ffn}
    \mathrm{FFN}(x) = \max \left(0, x W_{1}+b_{1}\right) W_{2}+b_{2},
\end{align}
where $x$ denotes the input of FFN which can be the output hidden state of Equation~\ref{equ:self-attn} for each word.
To sum up, the encoder in the Transformer consists of multiple identical layers with multi-head self-attention (Equation~\ref{equ:self-attn}) and FFN layer (Equation~\ref{equ:ffn}), and we use the operator $\text{Enc}$ to denote this procedure:
\begin{align}\label{equ:trans-enc}
    \{{\bf h}_{0}, {\bf h}_{1}, \cdots, {\bf h}_{L_r}\} = \text{Enc}({\text{[CLS]}, r_1, \cdots, r_{L_r}}),
\end{align}
where the output is the hidden states of each token in the input text $r$, and $\text{[CLS]}$ is a special token inserted at the start of the input text.
The hidden state ${\bf h}_{0}$ of the special token $\text{[CLS]}$ can aggregate information from all the tokens~\cite{Liu2019RoBERTaAR,Devlin2019BERTPO,Gao2022HeteroQA}.

In the decoder module, we first apply a multi-head self-attention layer on the mask output text embeddings which prevents attending to subsequent positions~\cite{Vaswani2017Attention}.
Next, we modify the self-attention (shown in Equation~\ref{equ:self-attn}) as the cross-attention layer which uses the current decoding state as the query $(e(r) W^Q)$ and uses the hidden states of input review as key and value.
Then, we use the FFN layer and linear projection layer with the softmax function to predict the word distribution of the generated text.
To increase the representation ability of the transformer framework, researchers usually employ multiple encoder and decoder layers.

\subsection{Pre-train Language Model}

Recently, large-scale language models based on transformer have been explored further advanced the state-of-the-art on many language understanding~\cite{Zhang2021UNBERT,Gu2021An,Song2021BoB,Gu2021Partner,Gu2021DialogBERT} and generation tasks~\cite{Zhang2019DialoGPT,Feng2021Language}.
These methods usually pre-train the transformer framework with mask language modeling~\cite{Devlin2019BERTPO,Yang2019XLNet} or text infilling~\cite{Lewis2020BART} task on large-scale text datasets.
In this paper, we employ the pre-train language model BART~\cite{Lewis2020BART} as the backbone of our review summary generation model, which can increase the fluency of the generated summary.
Although these pre-trained language models provide superior text generation ability, these methods usually use plain text as input and they cannot fully utilize the structure and manifold contextual information.
Next, we will introduce how to fine-tune a language model to generate a better review summary by incorporating historical customer and product reviews.

\section{\model Model}\label{sec:model}

\begin{figure*}
    \centering
    \includegraphics[width=1.6\columnwidth]{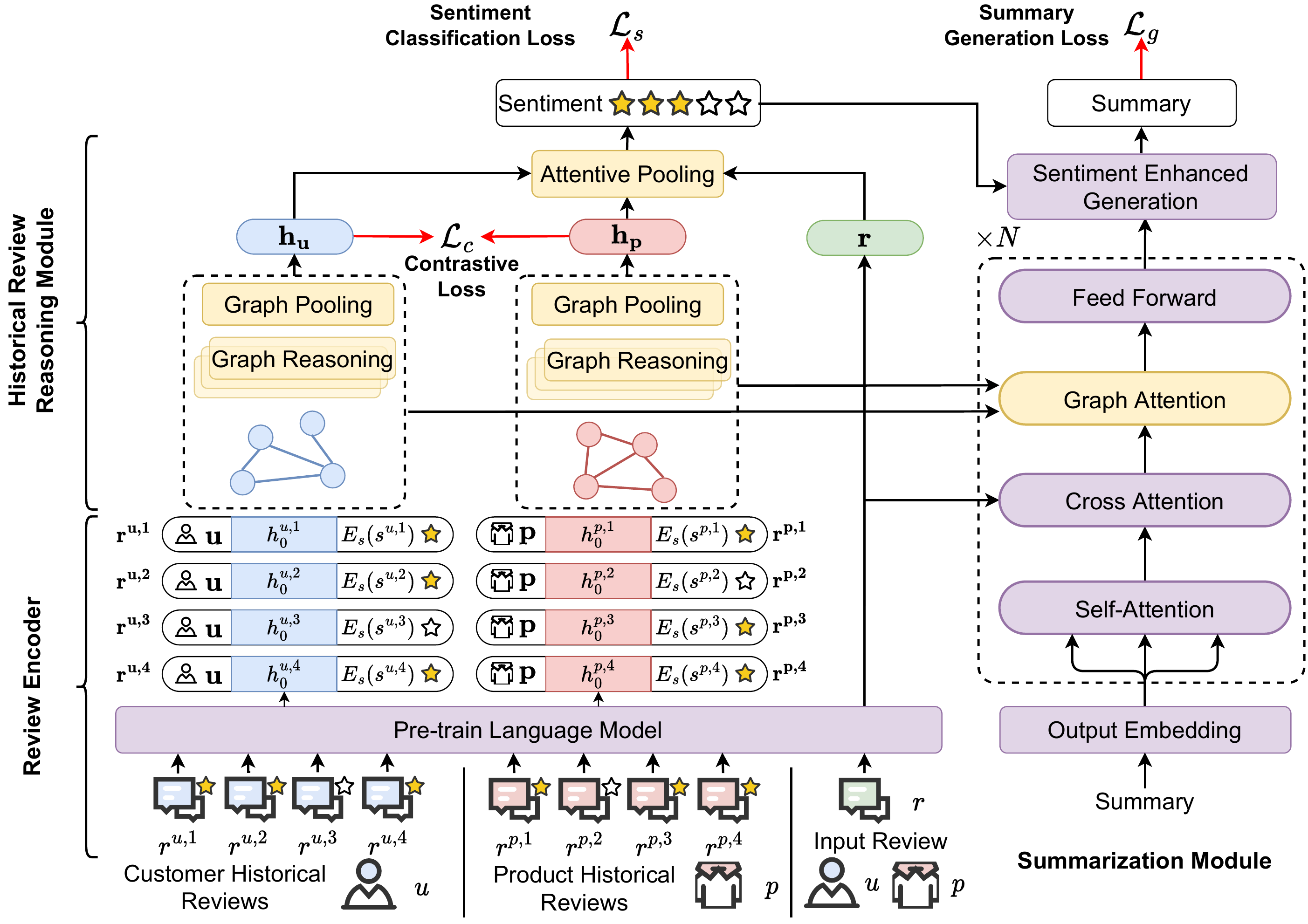}
    \caption{
        Overview of \model. Our model can be divided into four parts: (1) \textit{Review Encoder} encodes the review text into a vector and concatenate the customer or product representation with rating embedding; (2) \textit{Historical Review Relationship Encoder} constructs graphs for two types of reviews and conduct reasoning on these graphs; (3) \textit{Sentiment Classification Module} predicts the sentiment of input review by incorporating the reasoning result, and introduces a contrastive learning objective; (4) \textit{Summarization Module} generates the review summary.
    }
    \label{fig:model}
\end{figure*}

\subsection{Overview}

In this section, we introduce the \fullmodel (\model).
An overview of \model is shown in Figure~\ref{fig:model}, which has four main parts:

\noindent $\bullet$ \textbf{Review Encoder} encodes the review text into vector representation.

\noindent $\bullet$ \textbf{Historical Review Reasoning Module} constructs the relationship of product and customer reviews separately and employs a graph model to conduct reasoning. 

\noindent $\bullet$ \textbf{Sentiment Classification Module} incorporates graph representations for historical reviews to predict the rating for the input review. In order to force the model to learn heterogeneous information from two graphs, we employ a contrastive learning objective.

\noindent $\bullet$ \textbf{Summarization Module} first fuses the graph representations with the input review and then generates the summary,

\subsection{Review Encoder}\label{sec:review-enc}

To encode the reviews into vectors, we employ a pre-trained language model as the encoder:
\begin{equation}\label{equ:utterance-encoder}
    \{{\bf h}^{*}_{0}, {\bf h}^{*}_{1}, \cdots, {\bf h}^{*}_{L_r}\} = \text{Enc}({\text{[CLS]}, r^{*}_1, \cdots, r^{*}_{L_r}}),
\end{equation}
where $\text{Enc}$ is the encoder (details in ~\S~\ref{sec:preliminary}) in BART which outputs the vector ${\bf h}^{*}_{i} \in \mathbb{R}^d$ of $i$-th word $r^{*}_i$ in review $r^{*}$, and $r^{*}$ can be input review $r$, product review $r^{p,k}$ and customer review $r^{u,k}$.  %
To obtain an overall representation of the review $r^{*}$, we extract the hidden state ${\bf h}^{*}_{0}$ of the input special token $\text{[CLS]}$ as the representation $\mathbf{\hat{r^{*}}} = {\bf h}^{*}_{0}$.

Since the rating for historical customer reviews reflects the rating preference of the customer and the rating for the product reviews indicate the common sentiment for the customers who have already bought it, we propose to incorporate the rating into the review representation.
Thus, we first introduce an embedding matrix $E_s \in \mathbb{R}^{5 \times d}$ for each rating score (1-5) and combine the ratings for reviews into the review representation (shown in Equation~\ref{equ:review-repre}).

The product review written is highly associated with the customer's preference and the product attributes.
To better understand the review, we propose to use customer embedding to store personal preference information.
We use the embedding $\mathbf{u} \in \mathbb{R}^d$ as the representation for a customer of review $r^{u,k}$.
Similarly, we also employ a product embedding $\mathbf{p} \in \mathbb{R}^d$ for the product $p$ of review $r^{p,k}$.
The user embedding and product embedding are all trainable parameters that are jointly optimized when training the model.
Finally, we combine the previous information as the final review representation:
\begin{align}\label{equ:review-repre}
    \mathbf{r^{u,k}} &= {\bf h}^{u,k}_{0} + \mathbf{u} + E_s(s^{u,k}),\\
    \mathbf{r^{p,k}} &= {\bf h}^{p,k}_{0} + \mathbf{p} + E_s(s^{p,k}), \\
    \mathbf{r} &= {\bf h}^{r}_{0},
\end{align}
where $\mathbf{r^{u,k}} \in \mathbb{R}^{d}$ is the representation for the $k$-th review written by the customer $u$, and $\mathbf{r^{p,k}} \in \mathbb{R}^{d}$ is the representation for the $k$-th review of product $p$. %

\subsection{Historical Review Reasoning Module}\label{sec:review-relation-model}

To model the relationship of product reviews and customer reviews separately, in this section, we propose to use a graph reasoning module.
First, we construct the review graph by using $2$ types of edge to for product $\mathcal{G}_p$ and customer $\mathcal{G}_u$ reviews:
\noindent $(1)$ \textbf{Time-aware Edge}: We first use the chronological relationship between reviews, which connects the review nodes according to the publish date. These relations can capture the dynamic rating tendency of users. 
\noindent $(2)$ \textbf{Rating-aware Edge}: Since the review with the same rating may share similar or related information, we also connect the review nodes with the same rating in each graph.

Next, we use the review vector representation (in Equation~\ref{equ:review-repre}) as the initial node representation.
After constructing the two graphs for product and customer reviews, we employ a \textbf{G}raph \textbf{C}onvolutional \textbf{N}etwork-based~\cite{Kipf2017SemiSupervisedCW,Tang2020Multihop,Cao2019Question} review reasoning module to conduct the message passing and reasoning between review nodes.
In this module, we apply the multi-layer graph convolution to aggregate information from neighbor nodes connected by the two type edges.
Since the different type edge represents different semantics, we should consider the edge type when passing information.
Inspired by \textbf{R}elational \textbf{G}raph \textbf{C}onvolutional \textbf{N}etwork (RGCN)~\cite{Schlichtkrull2018ModelingRD}, we employ a local information aggregation scheme, which iteratively updates the node representation based on immediate neighbors.
Different from GCN, RGCN propagates different information between nodes through the different type of relationships:
\begin{align*}
    h_{i}^{(l+1)} = \sigma\left(\sum_{q \in \mathcal{Q}} \sum_{j \in \mathcal{N}_{i}^{q}} \textstyle\frac{1}{\left| \mathcal{N}_{i}^{q} \right|} W_{r}^{(l)} h_{j}^{(l)} + W_{0}^{(l)} h_{i}^{(l)}\right),
\end{align*}
where $l$ denotes the layer index, $h_{j}^{(l)}, h_{i}^{(l)}$ are node representations, $\mathcal{N}_{i}^{q}$ denotes the node $i$'s neighbor nodes which are connected with relation $q$, $\mathcal{Q}$ is the relation type set contains two type of relations, $W_{r}^{(l)}, W_{0}^{(l)}$ are all trainable parameters, and $\sigma$ is the activation function.
After applying $L$ layers iterative updating by RGCN, we can obtain the updated node representation for each node $\{h_{1}^{(L)}, \dots, h_{L_r}^{(L)}\}$.

Then, we employ a graph average pooling layer to combine the information from the graph nodes of customer and product reviews:
\begin{align}\label{equ:graph-pooling}
    \mathbf{h_{u}} &= \text{avg}\left(\{h_{u,1}^{(L)}, \dots, h_{u,L_r}^{(L)}\}\right), \\
    \mathbf{h_{p}} &= \text{avg}\left(\{h_{p,1}^{(L)}, \dots, h_{p,L_r}^{(L)}\}\right),
\end{align}
where $\text{avg}$ denotes the average graph pooling layer, and $\mathbf{h_{u}}$ and $\mathbf{h_{p}}$ are the graph representations for customer and product review graphs respectively.

Since we aim to extract the customer's personal information from the historical customer reviews and capture the main aspects of the product from the historical product reviews, we employ a contrastive training objective~\cite{Gao2021SimCSE,Chen2020A,Caron2020Unsupervised,Li2021Contrastive,Tian2020Contrastive} to prevent the model from learning homogeneous information from two graph modules.
Contrastive learning is an instance-wise discriminative approach that aims at making similar instances closer and dissimilar instances far from each other in representation space~\cite{He2020Momentum,Chen2021Wasserstein,Zhang2021Supporting,Yan2021ConSERT,Tong2021Directed,Jain2021Contrastive}.
Thus, this contrastive training objective encourages the graph reasoning module to learn different information from the historical customer reviews and product reviews for input review summarization.
To achieve better performance, it is important to design proper negative samples in contrastive learning.
Since our model is to extract different information from the historical customer and product reviews, for the product review reasoning module, we use the product review representation as the positive sample and use other customer review representations in the mini-batch as negative samples.
Our graph reasoning module is encouraged to learn a representation space where review representations from the same review type (\eg customer review or product review) are pulled closer and reviews from different review type are pushed apart.
Inspired by the recent progress~\cite{liang2021RDrop,Gao2021SimCSE,Lee2021Contrastive,Giorgi2021DeCLUTR} of applying contrastive learning on the text data which uses simple but efficient independently sampled dropout masks on the representation to produce the data augmentation, we also use the same dropout on the vector representation of two graph reasoning results $\mathbf{h_{u}}$ and $\mathbf{h_{p}}$: %
\begin{align}\label{equ:contrastive-dropout}
    \mathbf{\hat{h_{u}}} &= \text{Dropout}(\mathbf{h_{u}}), \\
    \mathbf{\hat{h_{p}}} &= \text{Dropout}(\mathbf{h_{p}}).
\end{align}
For training the customer review reasoning module, we use a similar training method that uses the customer review representation as the positive sample and use other product review representations in the mini-batch as negative samples.
Thus, we employ the contrastive loss function as an additional training objective:
\begin{equation}
    \begin{aligned}
        \mathcal{L}_c = \log \frac{e^{\operatorname{sim}\left(\mathbf{h_{u}}, \mathbf{\hat{h_{u}}}\right) / \tau}}{\sum_{j=1}^{B} e^{\operatorname{sim}\left(\mathbf{h_{u}}, \mathbf{\hat{h_{p}}}_j \right) / \tau}} + 
         \log \frac{e^{\operatorname{sim}\left(\mathbf{h_{p}}, \mathbf{\hat{h_{p}}}\right) / \tau}}{\sum_{j=1}^{B} e^{\operatorname{sim}\left(\mathbf{h_{p}}, \mathbf{\hat{h_{u}}}_j \right) / \tau}},
    \end{aligned} 
\end{equation}
where $B$ denotes the mini-batch, and the $\operatorname{sim}$ denotes the similarity function $\operatorname{sim}\left(a, b\right)=b^{\top} a / \tau$ and $\tau$ is the temperature.

\subsection{Sentiment Classification Module}

As shown in many previous studies~\cite{Ma2018A,Chan2020A}, jointly training the product review sentiment classification task with the review summarization task can boost the performance for both tasks.
In this paper, we also follow this paradigm to employ this multi-task setting.
However, previous studies~\cite{Ma2018A,Chan2020A} only use the input review itself when predicting the sentiment, the historical reviews of the customer contain the personal rating bias and the historical reviews provide the common focused aspect of the product.
Thus, we fuse the historical customer and product reviews with the input review together by an attentive pooling layer:
\begin{align}
    \hat{a} = \text{Softmax}\left( W_a [\mathbf{h_{u}} \oplus \mathbf{h_{p}} \oplus \mathbf{r}] + b_a \right) ,
\end{align}
where $W_a, b_a$ are all trainable parameters and $\hat{a} \in \mathbb{R}^3$.
Then, we apply a weighted sum operation by the attention score $\hat{a}$ and use a multilayer perceptron to predict the rating of the input review:
\begin{align}
    z &= \hat{a}_1\mathbf{h_{u}} + \hat{a}_2\mathbf{h_{p}} + \hat{a}_3\mathbf{r} \in \mathbb{R}^d , \label{equ:senti-cls-repre} \\
    \hat{s^r} &= \text{Softmax} (\text{MLP}(z)), \\
    \mathcal{L}_{s} &= -\frac{1}{B} \sum_{i}^B s^r \log (\hat{s^r}), \label{equ:senti-cls-loss}
\end{align}
where $\hat{s^r} \in \mathbb{R}^5$ is the predicted rating distribution for input review $r$ over $5$ rating class.
We employ the cross-entropy as the loss function $\mathcal{L}_{s}$ for this sentiment classification task.

\begin{table*}
\centering
\caption{Rouge score for summarization task. $\dagger$ means the results are referred from the original paper. All our ROUGE scores have a 95\% confidence interval of at most 0.24 as reported by the ROUGE.}
\small
\label{tab:main}
\resizebox{0.7\linewidth}{!}{
     \begin{tabular}{c |c| c cc|cc c|cc c}
          \bottomrule
          \multirow{2}{4em}{Dataset}&\multirow{2}{3em}{System} &\multicolumn{3}{c}{Rouge-1} & \multicolumn{3}{c}{Rouge-2} &\multicolumn{3}{c}{Rouge-L} \\
          \cline{3-11}
          &&R&P&F&R&P&F&R&P&F \\
          \hline
          \hline
          
     \multirow{11}{4em}{Home} &\texttt{PGNet}~\cite{See2017Get} $\dagger$
          & 14.82 & 20.53 & 16.44 & 6.28 & 8.83 & 6.90 & 14.64 & 20.23 & 16.23 \\
     
          &\texttt{Max+copy}~\cite{Ma2018A} $\dagger$
          & 14.92 & 20.57 & 16.52 & 6.33 & 8.84 & 6.94 & 14.72 & 20.25 & 16.30 \\

          &\texttt{HSSC+copy}~\cite{Ma2018A} $\dagger$
          & 14.93 & 20.62 & 16.54 & 6.34 & 8.87 & 6.95 & 14.75 & 20.32 & 16.33 \\
          
          &\texttt{C.Transformer}~\cite{Gehrmann2018BottomUp} $\dagger$
          & 13.75 & 19.35 & 15.36 & 5.44 & 7.70 & 6.01 & 13.58 & 19.06 & 15.17\\
          
          &\texttt{DualView}~\cite{Chan2020A} $\dagger$
          & 15.18 & 20.96 & 16.81 & 6.57 & \textbf{9.19} & 7.21 & 15.00 & \textbf{20.65} & 16.60 \\

          &\texttt{TRNS}~\cite{Xu2021Transformer}
          & 13.92	&17.77	&14.60	&4.76&	5.82	&4.90	&13.66&	17.41&	14.32 \\
          
          &\texttt{Transformer}~\cite{Vaswani2017Attention}
           & 13.00	 &14.46&	13.69	&4.27&	4.58	&4.42&	12.75&	14.16	&13.42 \\
           
          &\texttt{BART}~\cite{Lewis2020BART} 
          & 20.09	&18.91	&19.48&	8.16	& 8.78	&8.46	&18.34&	19.62&	18.96 \\

          &\texttt{BART+Concat} 
          & 21.31	&19.03	&20.12 & 	9.02	& 8.33	&8.66	&19.62&	17.66&	18.59 \\
          
          &\texttt{BART+Senti}
          & 21.57&	16.23	&18.52&	9.09&	6.83&	7.80	&20.91&	15.79	&17.99 \\
          
          &\texttt{BART+Con.+Sen.}
          & 21.71 &18.40&19.92&	9.42&	7.83&	8.55& 21.01 & 18.28 & 18.97 \\

          &\model (Ours)
          & \textbf{22.06} &\textbf{19.09} &\textbf{20.47} &\textbf{9.83} &8.61 &\textbf{9.18} &\textbf{21.36} &18.54 &\textbf{19.85}
          \\
          
          \hline
          \hline
          
          \multirow{11}{4em}{Toys} & \texttt{PGNet}~\cite{See2017Get} $\dagger$
           &14.77 & 20.54 & 16.40 & 6.18 & 8.47 & 6.74 & 14.53 & 20.13 & 16.11  \\
     
          &\texttt{Max+copy}~\cite{Ma2018A} $\dagger$
          & 14.62 & 20.52 & 16.29 & 5.92 & 8.19 & 6.48 & 14.37 & 20.09 &15.99\\    
          
          &\texttt{HSSC+copy}~\cite{Ma2018A} $\dagger$
          & 14.70 & 20.29 & 16.27 & 6.18 & 8.38 & 6.71 & 14.46 & 19.88 &15.98\\
          
          &\texttt{C.Transformer}~\cite{Gehrmann2018BottomUp} $\dagger$
          & 12.57 & 17.53 & 13.94 & 4.76 & 6.42 & 5.14 & 12.36 & 17.16 & 13.69 \\
          
          &\texttt{DualView}~\cite{Chan2020A} $\dagger$
          & 14.83 & 20.76 & 16.50 & 6.17 & 8.57 & 6.75 & 14.57 & 20.30 & 16.19  \\

          &\texttt{TRNS}~\cite{Xu2021Transformer}
          & 21.18	&17.85	&19.37	&9.20&	7.22	&8.09	&21.26&	16.69&	18.70 \\
          
          &\texttt{Transformer}~\cite{Vaswani2017Attention}
           & 13.49 & 13.48 & 13.49 & 4.39 & 4.25 & 4.32 & 13.15 & 13.09 & 13.12 \\
           
           &\texttt{BART}~\cite{Lewis2020BART}
          & 21.05 & 16.25 & 18.34 & 9.01 & 6.83 & 7.77 & 20.75 & 15.71 & 17.88\\

          &\texttt{BART+Concat} 
          & 21.95	&17.79	&19.65&	9.22	& \textbf{8.24}	&8.70	&21.02&	17.67&	19.20 \\
          
          &\texttt{BART+Senti}
          & 21.55 & 16.28 & 18.55 & 9.06 & 6.47 & 7.55 & 20.99 & 15.73 & 17.98 \\
          
          &\texttt{BART+Con.+Sen.}
          & \textbf{23.00} &17.74&20.03&	9.14&	8.03&	8.55& 21.12 & 17.00 & 18.84 \\
          
     &\model (Ours)
          & 22.96 & \textbf{18.71} & \textbf{20.62} & \textbf{9.83} & 8.03 & \textbf{8.84} & \textbf{22.10} &\textbf{18.16} & \textbf{19.94} \\
          \hline
          \hline
          
          \multirow{11}{4em}{Sports}&\texttt{PGNet}~\cite{See2017Get} $\dagger$ 
          & 14.78 & 19.79 & 16.13 & 6.13 & 8.20 &6.62 & 14.58 & 19.46 & 15.89 \\
          
          &\texttt{Max+copy}~\cite{Ma2018A} $\dagger$
          & 14.75 & 19.86 & 16.15 & 6.11 & 8.22 & 6.62 & 14.56 & 19.53 & 15.92 \\

          &\texttt{HSSC+copy}~\cite{Ma2018A} $\dagger$
          &14.64 &19.61 &15.98&5.95&7.98&6.43&14.43&19.26&15.74 \\
          
          &\texttt{C.Transformer}~\cite{Gehrmann2018BottomUp} $\dagger$
          & 13.73 & 18.46 &15.02 &5.13 &6.88 &5.56 &13.54&18.13&14.80 \\
          
          &\texttt{DualView}~\cite{Chan2020A} $\dagger$
          & 15.39 & 20.53 & 16.79 & 6.46 & 8.63 & 6.98 & 15.18 & 20.19 & 16.55 \\

          &\texttt{TRNS}~\cite{Xu2021Transformer}
          & 12.35&	14.29&	13.25&	3.60	&3.97&	3.78	&12.17&	14.04	&13.04 \\
          
          &\texttt{Transformer}~\cite{Vaswani2017Attention}
           & 12.84 & 13.92 & 13.36 & 3.60 & 3.78 & 3.69 & 12.58 & 13.62 & 13.08\\
           
           &\texttt{BART}~\cite{Lewis2020BART}  
          & 21.33 & 17.31 & 19.11 & 9.20 & 7.51 & 8.27 & 20.66 & 16.80 & 18.53 \\

          &\texttt{BART+Concat} 
          & 20.98	&17.15	&18.87&	9.19	& 7.31	&8.14	&20.26&	17.68&	18.88 \\
          
          &\texttt{BART+Senti}
          & \textbf{22.35}&	16.34 & 18.88&	 9.27&	6.69	& 7.77	& \textbf{21.57}& 	15.88	&18.29\\
          
          &\texttt{BART+Con.+Sen.}
          & 20.56 &14.67&19.12&	9.36&	7.39&	8.26& 21.26 & 16.79 & 18.76 \\
          
          &\model (Ours)
          & 21.61&	\textbf{18.82}	&\textbf{20.12}&\textbf{9.36}&	\textbf{8.01}&	\textbf{8.63}&	20.98	&\textbf{18.32}	&\textbf{19.56}\\
          
          \hline
          \hline

          \multirow{11}{4em}{Movies}&\texttt{PGNet}~\cite{See2017Get} $\dagger$ 
          & 12.67 & 17.76 & 14.04 & 5.14 & 7.38 & 5.66 & 12.40 & 17.32 & 13.72 \\
         
          &\texttt{Max+copy}~\cite{Ma2018A} $\dagger$
          & 12.61 & 17.81 & 14.01 & 5.04 & 7.32 & 5.57 & 12.34 & 17.38 & 13.69 \\    
          
          &\texttt{HSSC+copy}~\cite{Ma2018A} $\dagger$
          & 12.66 & 17.92 & 14.08 & 5.06 & 7.37 & 5.60 & 12.39 & 17.47 & 13.76\\
          
          &\texttt{C.Transformer}~\cite{Gehrmann2018BottomUp} $\dagger$
          & 12.09 & 16.78 & 13.34 & 4.46 & 6.30 & 4.89 & 11.81 & 16.33 & 13.01\\
          
          &\texttt{DualView}~\cite{Chan2020A} $\dagger$
          & 12.84 & 17.98 & 14.22 & 5.22 & 7.48 & 5.75 & 12.57 & 17.55 & 13.90 \\

          &\texttt{TRNS}~\cite{Xu2021Transformer}
          & 11.80 & 11.64 & 11.72 & 2.84 & 2.86 & 2.85 & 11.32 & 11.14 & 11.23 \\
          
          &\texttt{Transformer}~\cite{Vaswani2017Attention}
           & 13.44 & 17.00 & 15.01 & 5.26 & 6.69 & 5.89 & 12.81 & \textbf{18.12} & 15.01\\
           
           &\texttt{BART}~\cite{Lewis2020BART}  
          & 17.64 & 13.12 & 15.05 & 7.23 & 5.25 & 6.08 & 17.55 & 13.14 & 15.03\\

          &\texttt{BART+Concat} 
          & 18.56&15.72&17.02&7.35&7.08&7.21&18.01&14.92&16.32 \\
          
          &\texttt{BART+Senti}
          & 18.91 & 13.79 & 15.95 & 7.48 & 5.59 & 6.40 & 17.99 & 13.14 & 15.19 \\
          
          &\texttt{BART+Con.+Sen.}
          & 19.27 &17.14&18.14&	8.22&	7.00&	7.56& 18.54 & 15.51 & 16.89 \\
          
          &\model (Ours)
          & \textbf{20.56} & \textbf{17.66} & \textbf{19.00} & \textbf{9.16} & \textbf{8.05} & \textbf{8.57} & \textbf{19.74} & 16.99 & \textbf{18.26}\\
               
          \bottomrule
     \end{tabular}
}
\end{table*}

\subsection{Summarization Module}

Finally, to incorporate the two graph representations which capture the customer's personal information and the product-specific information in the generation process of the summary, we propose to modify the original transformer framework.
We first conduct the original self-attention and cross-attention layer in the transformer to incorporate the current decoded text and the input review $r$ respectively.
After these two layers, we obtain the hidden state $\mathbf{h}^d_t$ for decoding step $t$.
Next, we propose a graph-attention layer that extracts the useful knowledge from the nodes representation in customer review and product review graph:
\begin{equation}
    \begin{aligned}
        & \mathcal{H}^p_t = \operatorname{GraphAttn}(\mathbf{h}^d_t, \mathcal{G}_p) = \\
        & \operatorname{Softmax}\left(\frac{(\mathbf{h}^d_t W^Q) (\mathcal{G}_u W^K)}{\sqrt{d}}\right) (\mathcal{G}_p W^V),
    \end{aligned}
\end{equation}
where $\mathcal{G}_p = \{h_{p,1}^{(L)}, \dots, h_{p,L_r}^{(L)}\}$ is the set of graph node representations of thr product review graph, and $W^Q, W^K, W^V$ are all trainable parameters.
We conduct the same $\operatorname{GraphAttn}$ operator using different parameters on customer review graph nodes $\mathcal{G}_u$, and obtain the output hidden states $\mathcal{H}^u_t$.
Then, we combine the information from customer reviews $\mathcal{H}^u_t$ and product reviews $\mathcal{H}^p_t$ to obtain the hidden state for current decoding step:
\begin{align}
    \mathcal{H}^{\prime}_t &= \operatorname{MLP}(\mathcal{H}^u_t + \mathcal{H}^p_t),
\end{align}
where $\mathcal{H}^{\prime}$ is the combined information from both graphs.

Since the rating (sentiment) of the review can be seen as a high-level abstract of the review, the sentiment information can help the summarization module to capture the main idea of the review.
Thus, we propose the \textbf{sentiment enhanced generation} module which incorporates sentiment classification representation $z$ (calculated in Equation~\ref{equ:senti-cls-repre}) into final summary generation:
\begin{align}
    \delta = \text{Sigmoid}(W_{g1}\mathcal{H}^{\prime}_t + W_{g2}z + b_g), \label{equ:senti-aware-gen-gated}\\
    \mathcal{H}_t = \mathrm{FFN}(\mathcal{H}^{\prime}_t) + \delta z, \quad P^w_t = \operatorname{MLP}(\mathcal{H}_t),\label{equ:senti-aware-gen-hidden}
\end{align}
where $W_{g1}, W_{g2}, b_g$ are all trainable parameters, $P^w_t$ is the predicted token distribution for decoding step $t$.
The training objective is:
\begin{align}
    \mathcal{L}_g = \textstyle\sum_{t=0}^{L_r}-\log P^w_t\left(y_{t}\right).
\end{align}

Finally, we combine the training objectives for each module as the final training objective:
\begin{align}
    \mathcal{L} = \mathcal{L}_g + \mathcal{L}_s + \alpha\mathcal{L}_c,
\end{align}
where $\alpha$ is a hyper-parameter.
The gradient descent method is employed to update all the parameters in our model to minimize this loss function.

\newcommand{\cbkgrnd}{\cellcolor{blue!15}}
\newcommand{\sbbkgrnd}{\cellcolor{gray!65}}
\newcommand{\phantomtriangle}{\phantom{\dubbelop}}

\section{Experimental Setup}\label{sec:exp-setup}

\subsection{Dataset}\label{sec:dataset}

\begin{table}[t]
    \caption{Dataset statistics.}
    \label{tbl:data-statistics}
    \centering
    \resizebox{1.0\columnwidth}{!}{
    \begin{tabular}{lllll}
     \toprule
     & Sports & Movies & Toys & Home \\
     \midrule 
     \# of training samples & 183,714 & 1,200,601 & 104,296 & 367,395 \\
     \# of validation samples & 9,000 & 20,000  & 8,000 & 10,000 \\
     \# of test samples & 9,000 & 20,000 & 8,000 & 10,000 \\
     Avg. words of review & 108.3 & 167.1 & 125.9 & 120.9 \\
     Avg. words of summary & 6.7 & 6.6 & 6.8 & 6.8 \\
     \bottomrule
    \end{tabular}
    }
\end{table}

To validate the effectiveness of the proposed method, we conduct the experiments on Amazon review~\cite{McAuley2015ImageBasedRO}.
We adopt product reviews from the following four domains as our datasets: Sports, Movies, Toys, and Home.
In our experiments, each data sample consists of a review text, a summary, and a rating. 
We randomly split each dataset into training, validation, and testing sets. 
We list some basic statistics for this dataset in Table~\ref{tbl:data-statistics}.
We regard the rating of review as a sentiment label, which is an integer in the range of $[1, 5]$.

\subsection{Evaluation Metrics}

Following the previous review summarization works~\cite{Xu2021Transformer,Chan2020A}, we also use the word overlap-based \textbf{Rouge score}~\cite{lin2004rouge} as the evaluation metric for the summarization task. %
Due to the limited space, we only report the F-value of Rouge in other experiments.

Since only using automatic evaluation metrics can be misleading~\cite{Stent2005EvaluatingEM}, we also conduct the human evaluation by three well-educated Master students to judge 50 randomly sampled summaries.  %
The statistical significance of differences observed between the performance of two runs is tested using a two-tailed paired t-test and is denoted using \dubbelop\ (or \dubbelneer) for strong significance at $\alpha=0.01$.

For the sentiment classification, we use the \textbf{macro F1} (M.F1) and \textbf{balanced accuracy} (B.Acc)~\cite{Brodersen2010TheBA} as the evaluation metric which is widely used in text classification methods~\cite{Zhou2018Differentiated,Chan2020A}.
Since the rating of review is very imbalanced (\eg 58.0\% of reviews give rating 5 in Toys dataset), we employ the B.Acc which is a variant of the accuracy for imbalanced datasets~\cite{Brodersen2010TheBA,Kelleher2015FundamentalsOM}.

\subsection{Implementation Details}

We implement our experiments using PyTorch~\cite{Paszke2019PyTorchAI} based on the Transformers~\cite{wolf-etal-2020-transformers}.
We train our model on two NVIDIA V100 GPUs for one day. 
We employ the pre-trained BART-base model (with 6 layers for encoder and decoder, the number of attention head is 12, and the hidden size is 768) to initialize part of the parameters.
The hyper-parameter $\alpha$ is set to 0.1.

\subsection{Comparisons}

To prove the effectiveness of each module, we conduct ablation studies on Toys dataset, which removes each key module in \model, and then form $8$ baseline methods shown in Table~\ref{tab:ablations}.
Apart from the ablation study, we also compare with the following summarization baselines:

\noindent (1) \texttt{PGNet}~\cite{See2017Get} is an RNN-based abstractive summarization method with a copy mechanism.

\noindent (2) \texttt{Transformer}~\cite{Vaswani2017Attention} is an encoder-decoder structure based solely on the attention mechanism~\cite{Bahdanau2015NeuralMT}.

\noindent (3) \texttt{C.Transformer}~\cite{Gehrmann2018BottomUp} is a variant model of \texttt{Transformer} which equips with the copy mechanism. 

\noindent (4) \texttt{BART}~\cite{Lewis2020BART} is a pre-trained Transformer by using denoising mask language model as the training objective, and it has achieved SOTA performance on many text generation tasks.

\noindent (5) \texttt{BART+Concat} is a baseline method that we concatenate product and customer reviews into the input of \texttt{BART} to generate the summary.

\noindent (6) \texttt{BART+Senti} is an intuitive baseline method that we use the \texttt{BART} to generate the summary and use the encoder hidden state to predict the review sentiment as an auxiliary task.

\noindent (7) \texttt{BART+Con.+Sen.} adds the sentiment classification task to the \texttt{BART+Concat}.

\noindent (8) \texttt{HSSC+Copy}~\cite{Ma2018A} is a review summarization model for jointly improving review summarization and sentiment classification with copy mechanism~\cite{See2017Get}.

\noindent (9) \texttt{Max+Copy}~\cite{Ma2018A} A bi-directional gated recurrent unit~\cite{Chung2014EmpiricalEO} based sequence-to-sequence architecture with copy mechanism and it uses the hidden states of the encoder to predict the review sentiment.

\noindent (10) \texttt{DualView}~\cite{Chan2020A} is a dual-view model that jointly improves the performance of the review summarization task and sentiment classification task.

\noindent (11) \texttt{TRNS}~\cite{Xu2021Transformer} propose the state-of-the-art transformer-based reasoning framework for personalized review summarization.

We also employ a strong sentiment classification method \texttt{DARLM}~\cite{Zhou2018Differentiated} and fine-tune the \texttt{BERT}~\cite{Devlin2019BERTPO} on the sentiment classification task.

\begin{table*}
    \centering
    \caption{Sentiment classification results. $\dagger$ means the results are referred from the original paper.}
    \label{tab:rating}
    \resizebox{1.1\columnwidth}{!}{
         \begin{tabular}{c |c c| cc|cc| cc}
              \bottomrule
         
              \multirow{2}{4em}{System} & \multicolumn{2}{c}{Movies} &\multicolumn{2}{c}{Toys}&\multicolumn{2}{c}{Sports}&\multicolumn{2}{c}{Home} \\
         
              \cline{2-9}
              & M.F1 & B.Acc & M.F1 & B.Acc & M.F1 & B.Acc & M.F1 & B.Acc \\
              \hline
              \hline
              \multicolumn{9}{@{}l}{\emph{Joinly Training for Review Summarization \& Sentiment classification}} \\
              \texttt{Max+copy} $\dagger$
              &60.67&59.23&54.24&53.66&53.27&51.99&58.51&57.42\\
              \texttt{HSSC+copy} $\dagger$
              &60.69&59.32&54.38&53.32&53.14&52.63&58.78&58.02\\
              \texttt{DualView} $\dagger$
              &62.00&60.52&55.70&54.06&56.31&54.28&60.73&59.63\\
              \texttt{BART+Senti}
              &61.99&61.13&58.88&57.12&58.63&57.17&61.88&61.23\\
    
              \hline
              \multicolumn{9}{@{}l}{\emph{Sentiment classification Only}} \\
              \texttt{DARLM} $\dagger$
              &57.75&53.96&50.58&48.67&49.60&47.95&54.49&53.43\\
              \texttt{BERT}
              & 59.82&	58.98& 57.42&	56.32&56.49	&55.38& 58.97&	58.6 \\
              \texttt{Roberta}
              & 60.87&	60.13& 58.04	&57.27&\textbf{61.41}	&\textbf{60.93}& 61.41&	60.93 \\
              
              \hline
    
              \model (Our)
              & \textbf{63.42}&	\textbf{61.99}& \textbf{61.56}&	\textbf{61.08}&60.58	&58.87& \textbf{62.46}&	\textbf{62.26} \\
              \bottomrule
         \end{tabular}
    }
\end{table*}

\begin{table}[t]
    \centering
    \caption{Ablation models for comparison.}
    \label{tab:ablations}
    \resizebox{0.9\columnwidth}{!}{
    \begin{tabular}{@{}l|l}
        \toprule
        Acronym & Gloss \\
        \hline
        \hline
        \model-CR &  \multicolumn{1}{p{6cm}}{w/o \textbf{C}ustomer \textbf{R}eviews}\\
        \model-PR &  \multicolumn{1}{p{6cm}}{w/o \textbf{P}roduct \textbf{R}eviews}\\
        \model-MIX &  \multicolumn{1}{p{6cm}}{w/ \textbf{MIX}ed customer and product reviews}\\
        \model-CL &  \multicolumn{1}{p{6cm}}{w/o \textbf{C}ontrastive \textbf{L}oss}\\
        \model-SC &  \multicolumn{1}{p{6cm}}{w/o \textbf{S}entiment \textbf{C}lassification Loss (Eq.~\ref{equ:senti-cls-loss})}\\
        \model-SEG &  \multicolumn{1}{p{6cm}}{w/o \textbf{S}entiment-\textbf{E}nhanced \textbf{G}eneration (Eq.~\ref{equ:senti-aware-gen-gated})}\\
        \model-HR &  \multicolumn{1}{p{6cm}}{w/o \textbf{H}istorical \textbf{R}ating}\\
        \model-Graph &  \multicolumn{1}{p{6cm}}{Remove the \textbf{Graph} module and generator attends to original historical reviews}\\
        \bottomrule
    \end{tabular}
    }
\end{table}

\section{Experimental Results} \label{sec:exp-result}

\subsection{Overall Performance}\label{sec:overall-exp}

We compare our model with the baselines listed in Table~\ref{tab:main}.
Our model performs consistently better on four datasets than other state-of-the-art review summarization models with improvements of 10.53\%, 17.69\%, and 10.34\% on the Home dataset, and achieves 11.16\%, 17.09\%, and 10.90\% improvements on the Toys dataset compared with \texttt{BART+Senti} in terms of F-value of Rouge-1, Rouge-2, and Rouge-L respectively.
This demonstrates that our method achieves better performance than previous strong baselines not only because we use a pre-trained language model, but also because we use model historical customer and product reviews separately and incorporate the historical rating information.

\begin{table}
    \begin{center}
    \caption{Human evaluation results on Toys dataset.}
    \label{tab:human_eval}
    \resizebox{0.7\columnwidth}{!}{
    \begin{tabular}{c|ccc}
    \toprule
    Method & Fluency & Informativeness & Factuality \\
    \hline
    \hline
    \texttt{BART+Senti}  & 2.18 & 1.92 & 2.14 \\ 
    \texttt{BART}  & 2.02 & 2.02 & 1.91 \\
    \cbkgrnd \texttt{DualView} & \cbkgrnd 2.10 & \cbkgrnd 1.98 & \cbkgrnd 2.16 \\
    \midrule
    \model     & \textbf{2.32}\dubbelop & \textbf{2.34}\dubbelop & \textbf{2.28}\dubbelop \\ 
    \bottomrule
    \end{tabular}
    }
    \end{center}
\end{table}

For the human evaluation, we asked the annotators to rate the generated summary according to its fluency, informativeness, and factuality on the Toys dataset.
The rating score ranges from 1 to 3, with 3 being the best.
Table~\ref{tab:human_eval} lists the average scores, showing that \model outperforms the other baseline models in terms of fluency, informativeness, and factuality.
The kappa statistics are 0.44, 0.52, and 0.43 for fluency, informativeness, and factuality, and that indicates moderate agreement between annotators.
We also conduct the paired student t-test between \model and \texttt{DualView} and obtain $p < 0.05$ for all metrics.
From this experiment, we find that the \model outperforms the baselines in all metrics, which demonstrates the \model can generate fluent summaries with correct facts.

For the sentiment classification task, from Table~\ref{tab:rating}, we can find that our model also achieves superior performance among most of the review sentiment classification methods and state-of-the-art methods.

\subsection{Ablation Studies}\label{sec:ablation-exp}

\begin{table}
    \centering
    \caption{Ablation study on the Toys dataset.}
    \label{tab:ablation-exp}
    \resizebox{0.9\columnwidth}{!}{
         \begin{tabular}{c |c| c| c|c|c}
              \bottomrule
         
              Model & Rouge-1 & Rouge-2 & Rouge-L  &B.Acc& M.F1\\
         
              \hline
              \hline
                 
              \model
              &\textbf{20.62}&8.84&\textbf{19.94}&61.08&\textbf{61.56}\\
              
              \model-CR
              &19.55&8.29&19.01&60.24&60.05\\
              
              \model-PR
              &19.62&8.12&19.22&\textbf{61.13}&60.15 \\
              
              \model-MIX
              &20.00&7.92&18.43&60.02&59.86\\
              
              \model-CL
               &19.40&\textbf{8.92}&19.55&59.71&60.59\\
            
              \model-SC
              &19.53&7.51&19.06&-&- \\
    
              \model-SEG
              & 19.89 & 7.99 & 18.45 & 59.27 & 59.55 \\
              
              \model-HR
              &19.63&8.45&19.39&59.12&60.01\\

              \model-Graph
              &18.86&7.94&19.01&59.82&59.26\\
              \bottomrule
         \end{tabular}
    }
\end{table}

\begin{table}
    \centering
    \caption{Influence of using different graph edges. Experiments are conducted on Toys dataset.}
    \label{tab:graph-edge}
    \resizebox{1.0\columnwidth}{!}{
         \begin{tabular}{c |c| c| c|c|c}
              \bottomrule
              Graph Construction & Rouge-1 & Rouge-2 & Rouge-L &B.Acc & M.F1\\
              \hline
              \hline
              \model
              &\textbf{20.62}&\textbf{8.84}&\textbf{19.94}&\textbf{61.08}&\textbf{61.56}\\
              w/o time-aware
              &19.58&8.21&19.31&60.71&61.40 \\
              w/o rating-aware
              &20.01&8.51&18.99&59.87&60.12 \\
              \bottomrule
         \end{tabular}
    }
\end{table}

\begin{figure*}
    \centering
    \includegraphics[width=2.0\columnwidth]{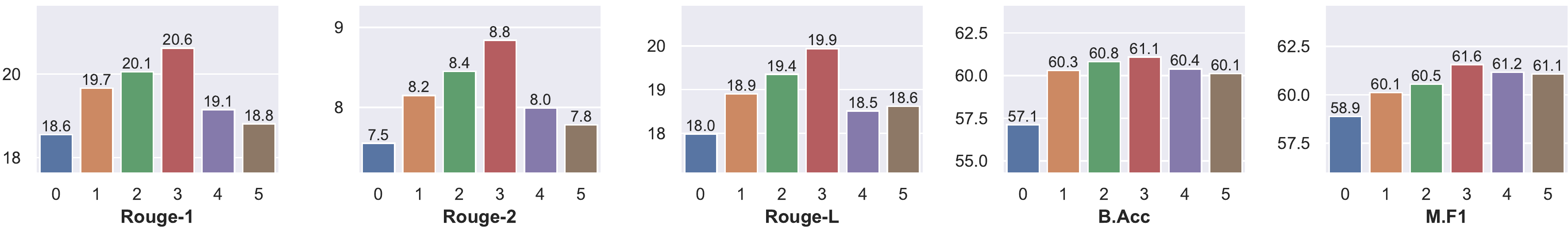}
    \caption{
        Influence of historical reviews number.
    }
    \label{fig:review-nums2}
\end{figure*}

We report the Rouge F-value of ablation models in Table~\ref{tab:ablation-exp}.
Most ablation models perform worse than \model, which demonstrates the preeminence of each module.
As proven by previous research work~\cite{Chan2020A}, jointly training the review sentiment classification and review summarization can boost the performance of both tasks, and our ablation models \model-SC and \model-SEG also verify this conclusion.

\textbf{Using two types of historical reviews}.
Ablation model \model-CR and \model-PR verify that only using the historical customer or product review cannot obtain good performance on both review summarization and sentiment classification tasks.
\model-CL performs worse than \model, which proves our contrastive learning objective can help the model extract different useful information from two review sources.

\textbf{Separately modeling the two types of reviews by graph reasoning module}.
In this paper, we propose a novel graph-based review reasoning module to capture the relationship between historical reviews of customer and product separately in \S~\ref{sec:review-relation-model}, and the ablation model \model-MIX and \model-Graph verify the effectiveness of our review reasoning module.
When we replace the multi-layer graph module by directly attending to the BART-based review representations (\model-Graph), the Rouge-1 F-score decreases by 9.23\% compared to \model.

\textbf{Using rating information of historical reviews}.
One of our contributions is to incorporate the rating of historical reviews.
We concatenate the embedding of the review rating into the review representation in Equation~\ref{equ:review-repre}.
To verify the effectiveness of incorporating historical rating, we test the ablation model \model-HR which removes the rating embedding in Equation~\ref{equ:review-repre}.
Experimental results show that the performance decreases by 5.04\%, 2.84\%, and 2.60\% compared to \model in terms of Rouge-1, Rouge-L, and M.F1.

\subsection{Effectiveness of Review Relationship}\label{sec:graph-edge-exp}

In this paper, we propose to use two different review relationships in \S~\ref{sec:review-relation-model}: chronological and same rating relationships.
To verify the effectiveness of these two edges, we conduct two ablation models which only use one type of relationship (edge).
From the results shown in Table~\ref{tab:graph-edge}, we can find that the time-aware edge contributes most to the summarization and the rating-aware edge contributes most to the sentiment classification.
This phenomenon demonstrates that the historical rating is useful for sentiment classification.

\subsection{Discussion of Using Contrastive Learning}\label{sec:contrastive-exp}

\begin{table}
    \centering
    \caption{Contrastive dropout rate. Experiments are conducted on Toys dataset.}
    \label{tab:dropout}
    \resizebox{0.9\columnwidth}{!}{
         \begin{tabular}{c |c| c| c|c|c}
              \bottomrule
         
              Dropout Rate & Rouge-1 & Rouge-2 & Rouge-L &B.Acc & M.F1\\
         
              \hline
              \hline
            
              0.01 &20.54&8.58&19.85&29.92&29.70\\
              
              0.05 &20.31&8.53&19.58&32.18&30.65\\
              
              0.1 & 20.33 & 8.64& 19.61 & 59.4 & 59.8\\
              
              0.6 &\textbf{20.62}&\textbf{8.84}&\textbf{19.94}&\textbf{61.08}&\textbf{61.56} \\
              
              0.9 & 19.54 & 8.12 & 18.93 & 27.3 & 27.43 \\
              
              \bottomrule
         \end{tabular}
    }
\end{table}

In \model, we employ a contrastive learning constraint (Equation~\ref{equ:contrastive-dropout}) to encourage the two graph modules to learn heterogeneous information from customer reviews and product reviews separately.
We use a simple but efficient dropout mask to obtain the augmented data representations.
In this section, we investigate the performance influence of using different dropout rates on the mask.
Table~\ref{tab:dropout} shows the results for both tasks.
We can find that (1) using 60\% dropout rate on the data representation mask is the best choice; (2) \model can stably achieve strong performance on both tasks when the dropout rate is in a reasonable range $(0.01 \sim 0.9)$ without a big performance difference.

\subsection{Influence of Historical Review Numbers}\label{sec:review-num-exp}

Since our model requires multiple historical customer and product reviews to capture the customer's writing style and personal preference and the common focused aspect of the product, it is an intuitive research question how many historical reviews should be used in our model?
We conduct an experiment using a different number of historical reviews for customer and product respectively and show the result for both tasks in Figure~\ref{fig:review-nums2}, which is conducted on the Toys dataset.
From these results, we can find that using 3 historical reviews for the customer and product respectively can achieve the best performance for both tasks.

\subsection{Case Study}

\begin{table}[t]
\centering
\caption{Case study. \ding{72} denotes the rating of review.}
\label{tab:case}
\resizebox{1\columnwidth}{!}{
\begin{tabular}{l}
    \toprule
    \multicolumn{1}{p{1\columnwidth}}{\textbf{Product Description:} Scotch Gift Wrap Cutter is great for cutting gift wrap paper and curling ribbon.}\\
    \multicolumn{1}{p{1\columnwidth}}{\textbf{Customer historical reviews} : \ding{192} \ding{72}\ding{72}\ding{72}\ding{73}\ding{73} fun, but may get repetitive in short time; \ding{193} \ding{72}\ding{72}\ding{72}\ding{72}\ding{73} not any better than scissors but safer around kids'} \\
    \multicolumn{1}{p{1\columnwidth}}{\textbf{Product historical reviews}: \ding{192} \ding{72}\ding{72}\ding{72}\ding{72}\ding{72} i love it \& it 's not just for cutting gift wrap; \ding{193} \ding{72}\ding{72}\ding{72}\ding{72}\ding{72} great little cutter} \\
    \multicolumn{1}{p{1\columnwidth}}{\textbf{Input Review}: \ding{72}\ding{72}\ding{72}\ding{73}\ding{73} i love the idea of a quick and easy blade to cut with . however, i found it fairly hard to get the cut ``going''.  for any other applications than cutting a straight line -- which is not what this was designed for.} \\
    \midrule
    \multicolumn{1}{p{1\columnwidth}}{\texttt{BART+Senti} \textbf{Summary}: \ding{72}\ding{72}\ding{73}\ding{73}\ding{73} hard to get the cut ``going''} \\
    \multicolumn{1}{p{1\columnwidth}}{\texttt{DualView} \textbf{Summary}: \ding{72}\ding{72}\ding{72}\ding{73}\ding{73} hard to control.} \\
    \multicolumn{1}{p{1\columnwidth}}{\model \textbf{Summary}: \ding{72}\ding{72}\ding{72}\ding{73}\ding{73} cute idea, but hard to get the cut ``going''} \\
    \bottomrule
\end{tabular}
}
\end{table}

From the case shown in Table~\ref{tab:case}, although all the methods generate fluent answers, the facts described in \texttt{BART+Senti} and \texttt{DualView} are not comprehensive.
And \model describes both the positive and negative aspects, and this review writing style is the same as the customer's previous writing style.
\section{Conclusion}\label{sec:conclusion}

In this paper, we propose \fullmodel (\model) which incorporates the historical customer and product reviews with the rating information to generate a personalized review summary and predict the sentiment of the review.
In order to extract useful information from the historical reviews, we propose a graph-based reasoning module to capture the customer review preference and the commonly focused aspect of the product.
To encourage the model to learn different information from two types of reviews, we introduce a contrastive learning objective for the graph reasoning module.
Finally, we also propose a graph attention layer to dynamically incorporate the graph for generating a fluent summary.
Extensive experiments on four benchmark datasets demonstrate that \model outperforms state-of-the-art baselines in both review summarization and sentiment classification tasks.


\bibliographystyle{ACM-Reference-Format}
\bibliography{references-new}

\end{document}